\documentclass{article}
\usepackage{meta}
\usepackage{amssymb}
\usepackage{amsmath}
\usepackage{epsfig}
\usepackage{algorithm}
\usepackage{algpseudocode}

\title{Conditional Normalizing Flow Surrogate for Monte Carlo Prediction of Radiative Properties in Nanoparticle-Embedded Layers}

\makeatletter
\def\name#1{\gdef\@name{#1\\}}
\makeatother
\name{{\bf \large Fahime Seyedheydari$^1$, Kevin Conley$^2$, and Simo Särkkä$^3$}}

\address{
$^{1, 3}$Department of Electrical Engineering and Automation (EEA), Aalto University, Espoo, Finland \\
$^2$Department of Chemistry and Materials Science, Aalto University, Espoo, Finland \\
*corresponding author, E-mail: {\tt fahime.seyedheydari@aalto.fi}
}

\begin{document}
\maketitle

\begin{abstract}
We present a probabilistic, data-driven surrogate model for predicting the radiative properties of nanoparticle embedded scattering media. The model uses conditional normalizing flows, which learn the conditional distribution of optical outputs, including reflectance, absorbance, and transmittance, given input parameters such as the absorption coefficient, scattering coefficient, anisotropy factor, and particle size distribution. We generate training data using Monte Carlo radiative transfer simulations, with optical properties derived from Mie theory. Unlike conventional neural networks, the conditional normalizing flow model yields full posterior predictive distributions, enabling both accurate forecasts and principled uncertainty quantification. Our results demonstrate that this model achieves high predictive accuracy and reliable uncertainty estimates, establishing it as a powerful and efficient surrogate for radiative transfer simulations.
\end{abstract}

\section{Introduction}

Monte Carlo (MC) simulations, are a foundational tool in modeling radiative transfer through scattering and absorbing media, with widespread applications in thermal engineering~\cite{azarkish2019reliability}, biomedical optics~\cite{doronin2011online}, and in materials science~\cite{seyedheydari2022electromagnetic}. The Monte Carlo for Multi-Layered Media (MCML) algorithm by Wang et al.~\cite{wang1995mcml}, originally developed for biological tissues, has been extensively adapted for other types of layered media. 

In previous work~\cite{conley2021silica}, we applied the MC method to study the optical properties of a particle-containing layer at low volume fractions. This method involves launching a large number of photons at normal incidence into the layer, where they are either transmitted, reflected, or absorbed. Despite advances in GPU-based parallel processing, MC simulations remain computationally expensive for high-throughput applications. In this work, we combine the MC method with Mie theory to model radiation transport in layers with low particle volume fractions. 

While MC simulations are accurate and physically grounded, they remain computationally intensive, especially when repeated across a high-dimensional input space of wavelengths, particle sizes, and material properties. Despite the efficiency gains achieved through parallel processing on GPUs, large-scale or repeated simulations continue to be computationally prohibitive in high-throughput applications such as design optimization or inverse scattering problems.

To address this limitation, surrogate models have been introduced to approximate the forward mapping from physical parameters to optical responses, enabling significant cost reduction while preserving accuracy~\cite{lu2019efficient, kaya2018surrogate}. Fully connected neural networks (FCNNs), in particular, have demonstrated strong performance in learning deterministic relationships from Monte Carlo-generated data. With sufficient training data, neural networks can approximate complex functions and provide fast, accurate predictions~\cite{goodfellow2016deep}. Such networks have been successfully applied to surrogate modeling tasks, including the design of thin-film photovoltaic cells and other optical systems~\cite{kaya2018surrogate}.

Deterministic neural networks lack the capacity to quantify uncertainty—an essential feature for applications such as photonic material design and uncertainty-aware simulations~\cite{gal2016dropout}. We propose a probabilistic surrogate model based on Conditional Normalizing Flows (CNFs), which transform a simple base distribution (e.g., standard Gaussian) into a complex target distribution through invertible transformations~\cite{rezende2015variational, kobyzev2020normalizing,park2024solving}. By conditioning this transformation on physical input parameters (e.g., $\mu_a$, $\mu_s$, $g$, and $\rho$), CNFs learn the conditional distribution $p(\mathbf{x} \mid \mathbf{y})$, where $\mathbf{y}$ denotes the input parameters and $\mathbf{x}$ represents the radiative spectral outputs. To overcome the computational burden of radiative transfer in scattering media, we generated training data using Monte Carlo simulations executed on a high-performance server at Aalto University.

The proposed CNF framework enables the generation of diverse and physically consistent outputs for a given input, thereby capturing the inherent uncertainty in radiative transfer simulations. It produces posterior predictive distributions, allowing for both mean predictions and associated confidence intervals—an essential feature for uncertainty-aware modeling. By maintaining consistency with Monte Carlo-generated training data, the framework ensures physical interpretability and robustness. To validate its effectiveness, we train a RealNVP-based CNF model on MC-simulated datasets. The model successfully predicts full spectral profiles of reflectance, absorbance, and transmittance with high accuracy while also providing well-calibrated uncertainty estimates. Model performance is rigorously assessed through cross-validated posterior sampling and comparisons against deterministic neural network baselines.

\section{Methods}
The proposed surrogate modeling framework is structured around two main components: data generation via Monte Carlo simulations and probabilistic modeling using CNFs. We begin by providing an overview of this framework and its rationale. Subsequent subsections describe the data generation process, the architecture of the CNF model, the training objective, and the inference procedure in detail.

\subsection{Data Generation and Preprocessing}

To train the CNF model, we generated a comprehensive dataset using Monte Carlo radiative transfer simulations, specifically designed for layered media embedded with nanoparticles. Each simulation corresponds to a folder containing detailed input parameters and the resulting output spectra.

The input parameters include the wavelength-dependent absorption coefficient ($\mu_a$), scattering coefficient ($\mu_s$), anisotropy factor ($g$), and the particle size distribution function ($\rho$). These parameters represent the key physical and optical properties that influence light transport within the medium.

For each simulation, the Monte Carlo method outputs the spectral properties of the system, including total reflectance (computed as the sum of specular and diffuse components), absorbance, and transmittance.

\begin{figure}
\centering
\includegraphics[width=0.9\linewidth]{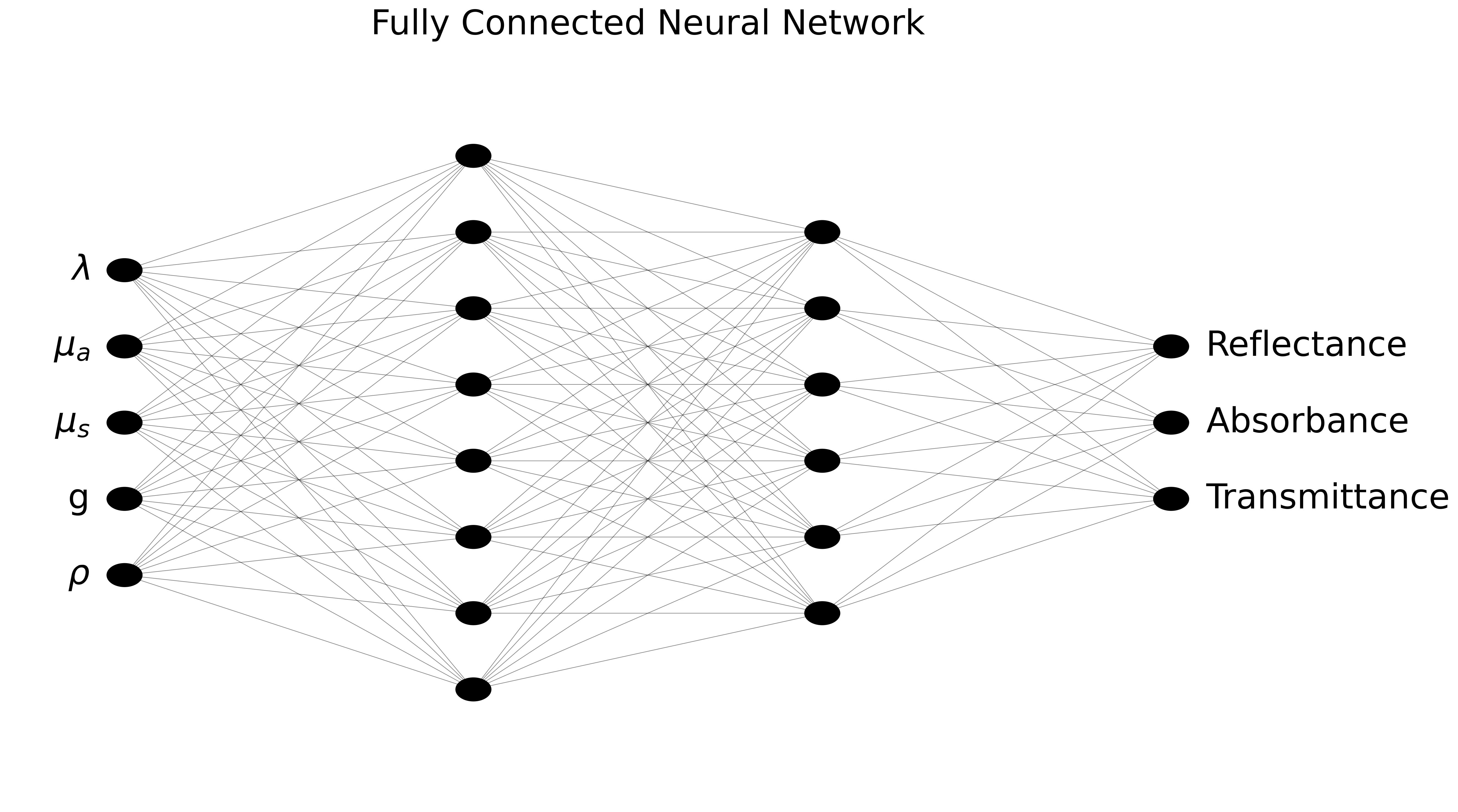}
\caption{A schematic representation of the fully connected neural network.}
\label{fig:schematic}
\end{figure}

All input features are preprocessed by stacking and reshaping them into a unified feature vector representing each sample. The corresponding outputs are vectorized in a similar manner, resulting in a structured dataset compatible with neural network training. A schematic overview of the network architecture used for modeling is provided in Figure~\ref{fig:schematic}.
\subsection{Conditional Normalizing Flow Architecture}

We employ a CNF framework to model the conditional distribution of radiative spectral outputs given input physical parameters. The CNF is based on a Real-valued Non-Volume Preserving (RealNVP) transformation, which provides a scalable and invertible mapping between a simple latent distribution and a complex target distribution~\cite{durkan2019neural}. In our case, the target distribution corresponds to spectral quantities such as total reflectance, absorbance, and transmittance, conditioned on high-dimensional input features extracted from the optical and material properties.

Starting from a standard Gaussian latent variable $\mathbf{z} \sim \mathcal{N}(\mathbf{0}, \mathbf{I})$, where $\mathbf{z} \in \mathbb{R}^D$ is a latent vector of the same dimensionality as the output and $\mathbf{I}$ is the identity covariance matrix, we define an affine mapping:
\[
\mathbf{x} = \exp\big(\log s(\mathbf{y})\big) \odot \mathbf{z} + t(\mathbf{y}),
\]
where $\log s(\mathbf{y})$ and $t(\mathbf{y})$ are the log-scale and shift vectors produced by a conditioner neural network, and $\odot$ denotes element-wise multiplication. This structure ensures that the transformation is differentiable and invertible, making it suitable for both sampling and likelihood estimation. Here, the functions $s(\mathbf{y})$ and $t(\mathbf{y})$ are parameterized by neural network weights $\psi$.

To compute the conditional likelihood $p(\mathbf{x} \mid \mathbf{y})$, we apply the change-of-variables formula. Due to the affine nature of the transformation, the Jacobian determinant simplifies to the sum of log-scale terms:
\[
\log p(\mathbf{x} \mid \mathbf{y}) = \log p(\mathbf{z}) + \sum_{j=1}^{D} \log s_j(\mathbf{y}),
\]
where $p(\mathbf{z})$ is the log-density under the standard normal prior. This formulation enables the model to learn expressive conditional distributions and supports principled uncertainty quantification via posterior sampling. For a full derivation of the conditional log-likelihood, see Supplementary Material, Section~\ref{Supplementary}.

\subsection{Training Objective}

The CNF model is trained to maximize the likelihood of the observed output data given the input features. This is achieved by modeling the conditional distribution $p(\mathbf{x} \mid \mathbf{y})$ using a bijective transformation applied to a latent variable sampled from a standard multivariate normal distribution. Specifically, we learn a mapping from latent space to data space through neural networks that output data-dependent shift and scale parameters.

Given a training pair $(\mathbf{y}^{(i)}, \mathbf{x}^{(i)})$, the model first computes the inverse transformation to recover the corresponding latent variable:
\[
\mathbf{z}^{(i)} = g_\psi^{-1}(\mathbf{x}^{(i)}, \mathbf{y}^{(i)}) = (\mathbf{x}^{(i)} - t(\mathbf{y}^{(i)})) \oslash \exp(\log s(\mathbf{y}^{(i)})),
\]
where $t(\mathbf{y})$ and $\log s(\mathbf{y})$ are the shift and log-scale vectors predicted by the conditioner neural network with parameters $\psi$, and $\oslash$ denotes element-wise division.

The conditional log-likelihood is then computed using the change-of-variables formula derived earlier:
\[
\log p_\psi(\mathbf{x}^{(i)} \mid \mathbf{y}^{(i)}) = \log p(\mathbf{z}^{(i)}) + \sum_{j=1}^{D} \log s_j(\mathbf{y}^{(i)}),
\]
where $\log p(\mathbf{z}^{(i)}) = -\frac{1}{2} \left( \|\mathbf{z}^{(i)}\|^2 + D \log(2\pi) \right)$, since $\mathbf{z}^{(i)} \sim \mathcal{N}(\mathbf{0}, \mathbf{I})$.

To train the model, we maximize the total log-likelihood over the dataset. Equivalently, the training objective is to minimize the negative log-likelihood loss:
\[
\mathcal{L}(\psi) = -\frac{1}{N} \sum_{i=1}^{N} \log p_\psi(\mathbf{x}^{(i)} \mid \mathbf{y}^{(i)}).
\]

This objective is minimized using stochastic gradient descent with the Adam optimizer. By learning parameters $\psi$ that maximize the likelihood of the observed data, the model captures the full conditional distribution and generates accurate, uncertainty-aware predictions of spectral responses.

\subsection{Training Procedure}

The model is trained using stochastic gradient descent with the Adam optimizer to minimize the negative log-likelihood objective. Training is conducted over 20{,}000 epochs using mini-batches. At the beginning of each epoch, the training data is shuffled to promote generalization and to avoid learning spurious patterns. For each mini-batch, the conditioner network computes the shift and log-scale parameters, defines the transformed distribution, evaluates the loss, and updates the model weights via backpropagation.

To ensure robust evaluation and reduce variance in performance estimates, we employ 5-fold cross-validation. The dataset is partitioned into five subsets, and training is repeated five times, each time using a different subset for validation and the remaining subsets for training. This yields five independently trained models. Model checkpoints are saved at the end of each fold to facilitate ensemble predictions and uncertainty estimation. The key steps of the training process are summarized in Algorithm~\ref{alg:cnf-training}.
\begin{algorithm}[H] 
\caption{Conditional Normalizing Flow Training}
\label{alg:cnf-training}
\begin{algorithmic}[1]
\Require Dataset $\{X, Y\}$, number of epochs $T$, batch size $B$
\Ensure Trained conditioner network

\State Initialize the conditioner network and Adam optimizer with learning rate $10^{-4}$
\For{$\text{epoch} = 1$ to $T$}
  \State Shuffle the training data $\{X, Y\}$
  \State Partition the data into mini-batches of size $B$
  \For{each mini-batch $(x_{\text{batch}}, y_{\text{batch}})$}
    \State {\footnotesize $(\texttt{shift}, \texttt{log\_scale}) \gets \texttt{conditioner\_net}(x_{\text{batch}})$}
    \State {\footnotesize $p(y_{\text{batch}} \mid x_{\text{batch}}) = \mathcal{N}(\texttt{shift}, \exp(\texttt{log\_scale}))$}
    \State $\mathcal{L} = -\log p(y_{\text{batch}} \mid x_{\text{batch}})$
    \State Update conditioner network via backpropagation
  \EndFor
\EndFor
\State \Return Trained conditioner network
\end{algorithmic}
\end{algorithm}
\subsection{Inference and Sampling}

After training, the CNF model is used to generate probabilistic predictions for new, unseen input configurations. Each input, represented by a high-dimensional vector of physical parameters, is passed through the conditioner network to compute the corresponding shift and log-scale vectors. These parameters define an affine transformation applied to latent samples drawn from a standard normal distribution, producing samples from the output distribution.

Specifically, given a new input \( \mathbf{y}_{\text{new}} \), the conditioner network outputs \( t(\mathbf{y}_{\text{new}}) \) and \( \log s(\mathbf{y}_{\text{new}}) \). A set of latent vectors \( \mathbf{z} \sim \mathcal{N}(\mathbf{0}, \mathbf{I}) \) is sampled and transformed into output predictions \( \mathbf{x} \) using the mapping:
\[
\mathbf{x} = \exp(\log s(\mathbf{y}_{\text{new}})) \odot \mathbf{z} + t(\mathbf{y}_{\text{new}}),
\]
where \( \odot \) denotes element-wise multiplication. This procedure is repeated across all models obtained from cross-validation, enabling ensemble sampling from the posterior predictive distribution.

The resulting samples are aggregated to compute the posterior mean and standard deviation for each spectral quantity. These statistics yield the predicted response along with its associated uncertainty. A 95\% confidence interval is computed as \( \hat{x}_{\text{mean}} \pm 1.96 \cdot \hat{\sigma} \), providing interpretable and calibrated probabilistic forecasts of total reflectance, absorbance, and transmittance for previously unseen physical inputs.

\subsection{Model Evaluation and Predictions}

Model evaluation is performed using the negative log-likelihood (NLL) computed on the validation sets during cross-validation. This metric quantifies how well the predicted conditional distributions align with the observed spectral responses and serves as a principled objective that is consistent with the training loss. Lower NLL values indicate higher model confidence and better agreement with the data.

To assess predictive performance, we compare the posterior means of the predicted outputs with the ground-truth spectral values across the wavelength domain. Additionally, the model provides standard deviations and confidence intervals, enabling uncertainty quantification for each prediction. These intervals offer valuable insights into the model’s reliability, particularly in the context of noisy or underdetermined physical systems.

In the following section, we present detailed comparisons between the predicted and true values, including graphical analyses with uncertainty bands. These evaluations underscore the model’s ability to generalize to unseen inputs and to provide well-calibrated probabilistic forecasts across a variety of optical configurations.

\section{Results and Discussion}

To evaluate the performance of our trained CNF model, we perform predictive inference on previously unseen data. The test input consists of wavelength-dependent optical properties—namely, the absorption coefficient ($\mu_a$), scattering coefficient ($\mu_s$), anisotropy factor ($g$), and particle size distribution ($\rho$)—which are concatenated into a single input vector and reshaped to match the input dimensionality expected by the trained model.

Predictions are generated using the ensemble of models obtained from 5-fold cross-validation. For each fold, the corresponding trained CNF model outputs shift and log-scale vectors conditioned on the test input. These parameters define a conditional Gaussian distribution from which we draw multiple samples (10,000 per model). The generated outputs are then transformed via the RealNVP bijector to yield predictive samples of the total reflectance, absorbance, and transmittance. The predictions from all five folds are aggregated to compute the posterior mean and standard deviation across the spectral domain, enabling robust estimation of both the predicted mean and associated uncertainty.

\begin{figure}[h]
\centering
\includegraphics[width=0.9\linewidth]{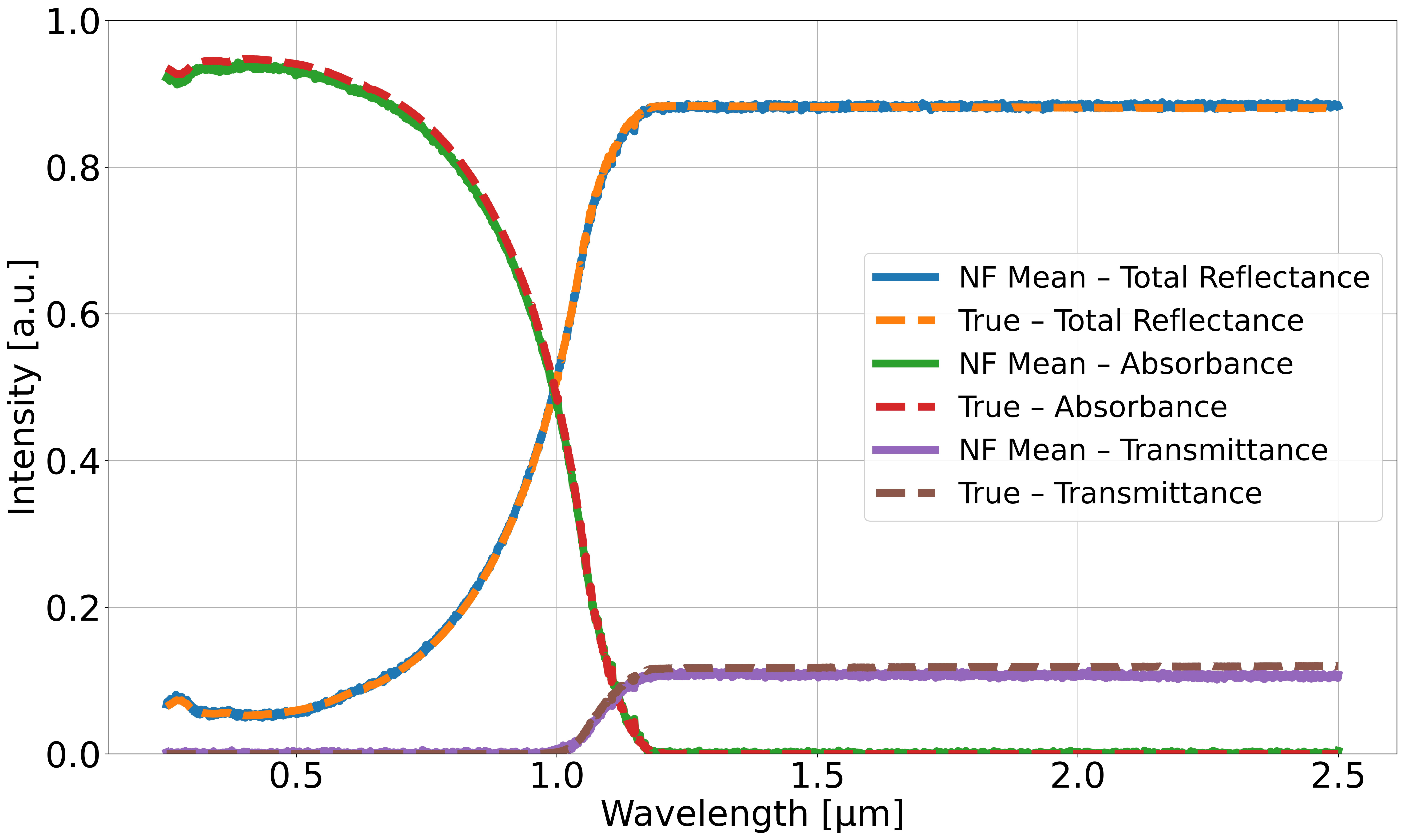}
\caption{Comparison between conditional Normalizing Flow model predictions (solid lines) and true simulation outputs (dashed lines) for total reflectance, absorbance, and transmittance across the spectral range.}
\label{fig:nf_vs_true_all}
\end{figure}

Figure~\ref{fig:nf_vs_true_all} presents a direct comparison between the NF-predicted mean spectra (solid lines) and the true Monte Carlo-generated values (dashed lines) for all three optical properties. The NF model achieves excellent agreement with the true values across the full wavelength range. Both reflectance and absorbance curves are accurately captured, including sharp transitions, while transmittance remains consistently modeled even in low-signal regions. This highlights the capacity of the CNF model to generalize across complex radiative behaviors in nanoparticle-laden media.
The high degree of overlap between predicted and true spectra confirms the accuracy of the CNF framework. The model remains stable in smooth spectral regions and accurately captures sharp spectral transitions. These results validate both the expressiveness of the CNF architecture and the robustness of the training strategy—including cross-validation and posterior sampling. Overall, the CNF model demonstrates strong generalization capabilities and reliability as a surrogate for fast, uncertainty-aware predictions in radiative transfer modeling.

\begin{figure}[h]
\centering
\includegraphics[width=0.9\linewidth]{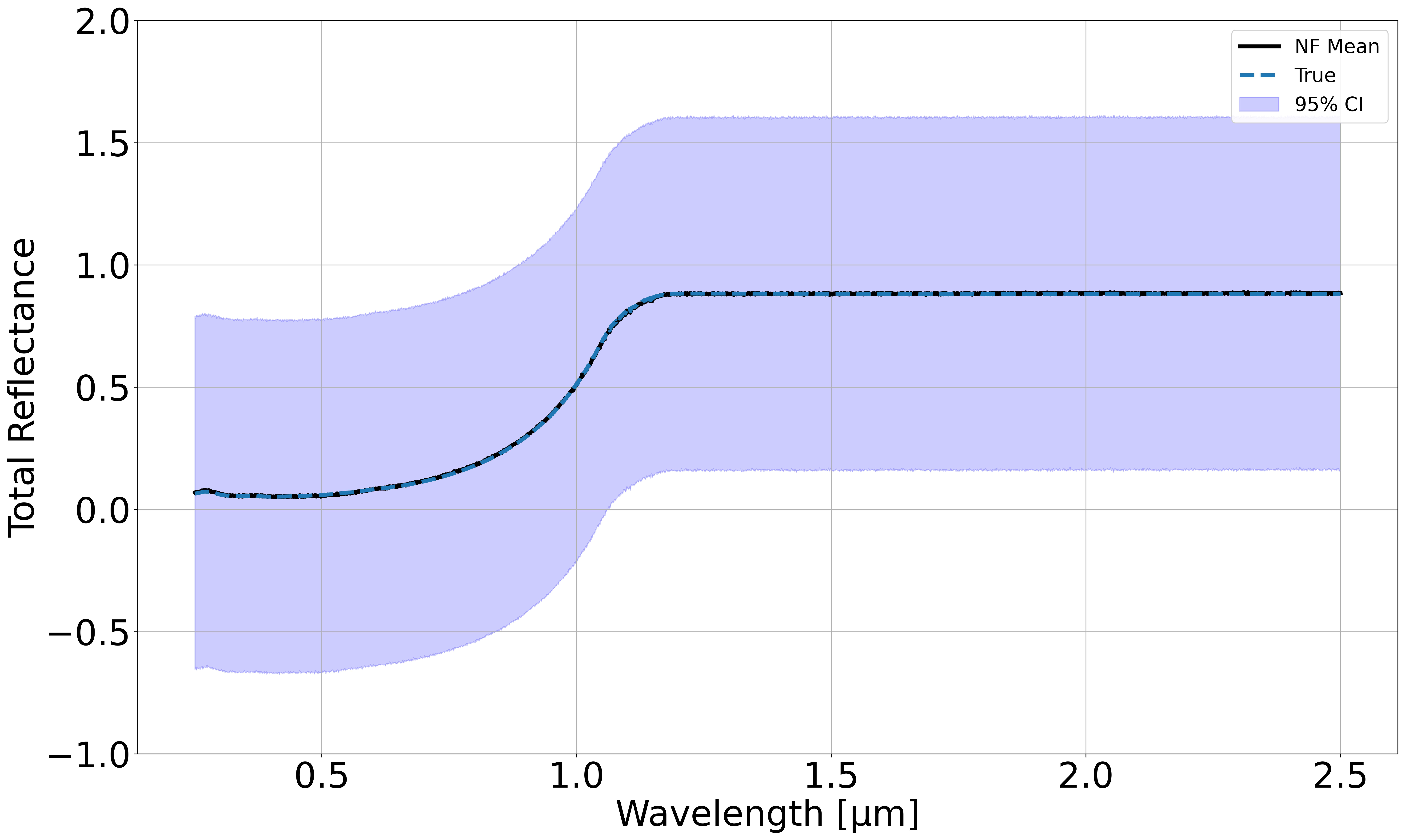}
\includegraphics[width=0.9\linewidth]{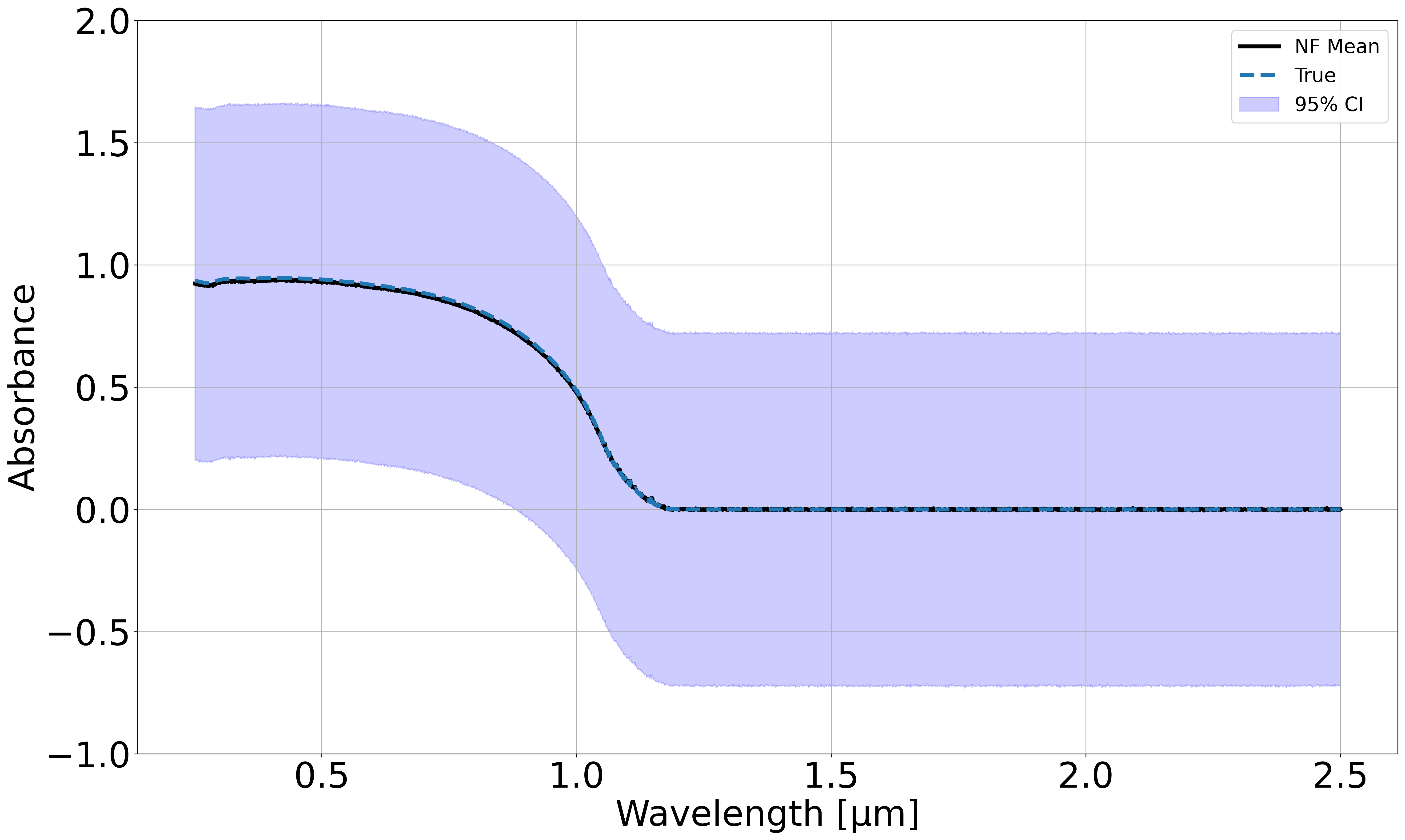}
\includegraphics[width=0.9\linewidth]{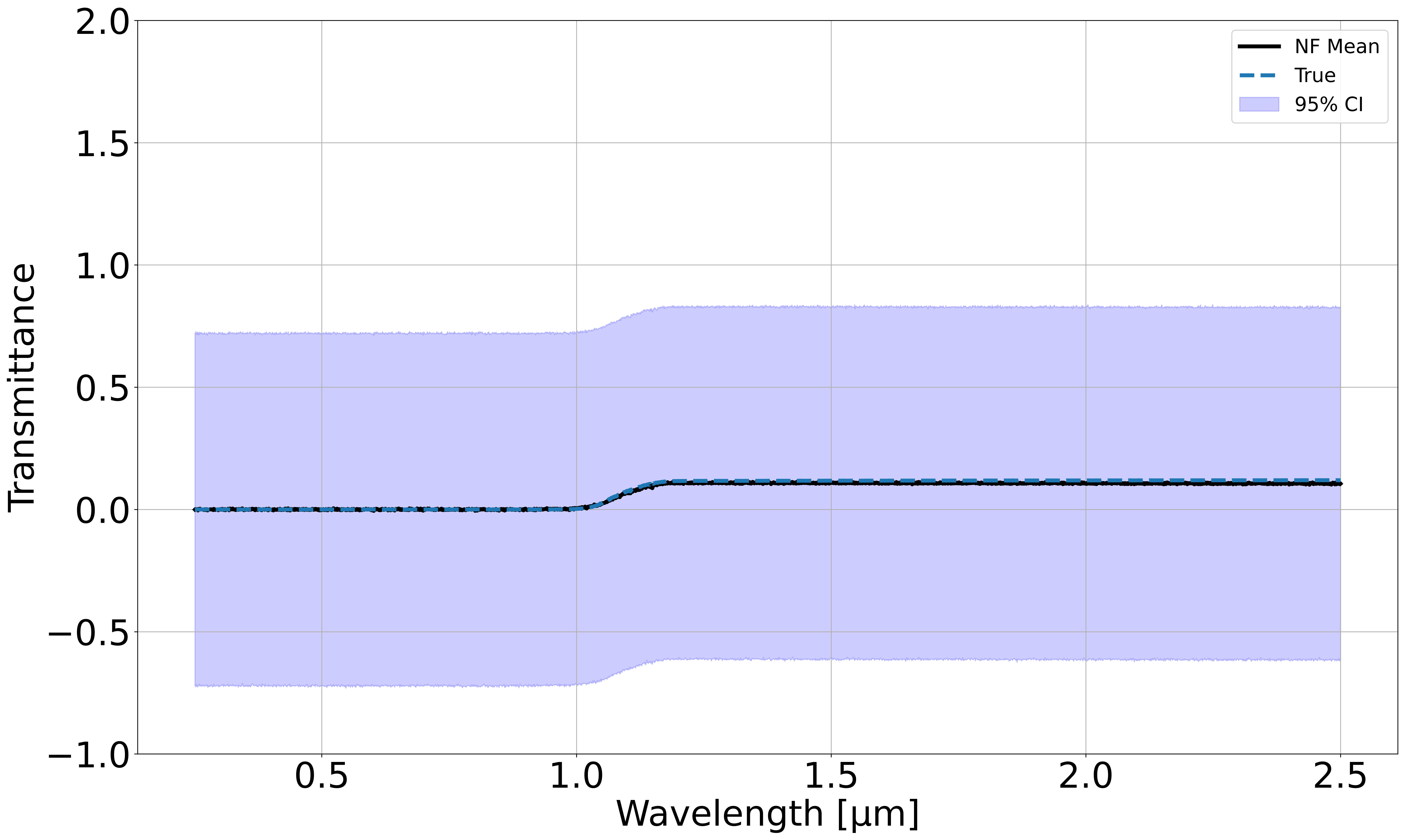}
\caption{Posterior predictions of (a) Total Reflectance, (b) Absorbance, and (c) Transmittance obtained from the CNF model. The solid black line indicates the predictive mean, the dashed blue line represents the true simulation output, and the shaded region denotes the 95\% confidence interval computed from posterior samples.}
\label{fig:combinedci}
\end{figure}

Figures~\ref{fig:combinedci} present individual comparisons between the CNF-predicted spectral properties and the corresponding ground truth values for (a) total reflectance, (b) absorbance, and (c) transmittance. Each plot displays the posterior predictive mean (solid black line), the true Monte Carlo simulation output (dashed blue line), and a 95\% confidence interval (shaded blue region) derived from posterior sampling. The close visual agreement between the predicted and true values highlights the model’s ability to faithfully capture spectral behavior across the entire wavelength range.

The confidence intervals remain relatively uniform across most of the spectrum, reflecting high predictive certainty. Even near the sharp spectral transition around 1.1~$\mu$m, the model maintains well-calibrated uncertainty, indicating robust learning and effective generalization in data-sparse or high-gradient regions. This behavior aligns with expected uncertainty dynamics in such scenarios. Notably, the true values consistently fall within the predicted bounds, further validating the reliability of the CNF model’s uncertainty estimates.

These findings confirm that the CNF model delivers both accurate point predictions and well-calibrated epistemic uncertainty quantification. Such reliable uncertainty estimates are crucial in optical modeling scenarios, where prediction confidence directly impacts experimental design, material selection, and downstream decisions. The strong performance demonstrated here suggests significant potential for extending this CNF-based approach to more complex forward and inverse problems in optical characterization.

While the conditional normalizing flow (CNF) model delivers accurate predictions and well-calibrated uncertainty estimates, it is worth noting that, in certain regions, the 95\% confidence intervals occasionally extend into negative values. Although mathematically permissible within the probabilistic framework of the model, such negative bounds are physically non-interpretable for radiative properties like reflectance, absorbance, and transmittance, which are inherently non-negative. This outcome highlights a known limitation of the current modeling approach. As part of our future work, we plan to enhance the model structure and incorporate additional constraints to ensure that both the predictive means and the corresponding confidence intervals remain consistent with physical reality.

\section{Conclusions}

This work introduced a probabilistic surrogate modeling framework based on Conditional Normalizing Flows (CNFs) to predict the spectral radiative properties—total reflectance, absorbance, and transmittance—of nanoparticle-embedded scattering media. Trained on synthetic data generated via Monte Carlo simulations combined with Mie theory-derived optical parameters, the model provides a fast, data-driven alternative to computationally expensive simulations.

The CNF framework successfully captures complex, wavelength-dependent optical behavior across a broad spectral range. Through 5-fold cross-validation and detailed uncertainty quantification, we demonstrated its ability to generalize to unseen inputs while producing well-calibrated posterior distributions. These predictive distributions enable interpretable confidence intervals and support robust, uncertainty-aware inference.

Our findings establish CNFs as powerful tools for forward modeling in optical and materials science. The probabilistic nature of the model is particularly advantageous in scenarios where predictive confidence is essential, including optical device design, inverse scattering, and experimental planning. Future extensions could explore support for more diverse particle geometries, higher concentrations, or multilayer systems—thereby broadening the relevance of the method in advanced photonics and nanomaterials research. 

Finally, we acknowledge that the model's confidence intervals can occasionally extend into negative values, which—although mathematically valid within the probabilistic framework—may be physically non-interpretable for radiative properties. Addressing this limitation is an important direction for future model development, ensuring that both predictive means and confidence intervals remain consistent with the underlying physics.

\section{Supplementary Material: Derivation of the Conditional Log-Likelihood in CNF}\label{Supplementary}
\textbf{Latent Variable and Forward Mapping:} Let $\mathbf{z} \in \mathbb{R}^D$ be a latent variable drawn from a standard multivariate Gaussian distribution:
\[
\mathbf{z} \sim \mathcal{N}(\mathbf{0}, \mathbf{I}),
\]
where $\mathbf{I} \in \mathbb{R}^{D \times D}$ is the identity matrix. We define an affine transformation to generate the observed variable $\mathbf{x} \in \mathbb{R}^D$ conditioned on an input $\mathbf{y}$:
\[
\mathbf{x} = \exp\big(\log s(\mathbf{y})\big) \odot \mathbf{z} + t(\mathbf{y}),
\]
where $\log s(\mathbf{y}) \in \mathbb{R}^D$ is the log-scale vector output by the conditioner network, $t(\mathbf{y}) \in \mathbb{R}^D$ is the shift vector from the same network, and $\odot$ denotes element-wise (Hadamard) product. This transformation is invertible and differentiable, making it suitable for both sampling and density estimation.

\textbf{Change of Variables Formula:} Given the invertible transformation $f_\psi: \mathbf{z} \mapsto \mathbf{x}$, the conditional log-probability $p(\mathbf{x} \mid \mathbf{y})$ is computed using the change-of-variables formula:
\[
\log p(\mathbf{x} \mid \mathbf{y}) = \log p(\mathbf{z}) - \log \left| \det \left( \frac{\partial \mathbf{x}}{\partial \mathbf{z}} \right) \right|.
\]

\textbf{Jacobian Determinant:}The Jacobian matrix $\frac{\partial \mathbf{x}}{\partial \mathbf{z}}$ is diagonal because the transformation is element-wise. Each diagonal entry is given by:
\[
\frac{\partial x_j}{\partial z_j} = \exp(\log s_j(\mathbf{y})) = s_j(\mathbf{y}).
\]
Hence, the determinant is:
\begin{align*}
\left| \det \left( \frac{\partial \mathbf{x}}{\partial \mathbf{z}} \right) \right| &= \prod_{j=1}^{D} s_j(\mathbf{y}) \\
\log \left| \det \left( \frac{\partial \mathbf{x}}{\partial \mathbf{z}} \right) \right| &= \sum_{j=1}^{D} \log s_j(\mathbf{y})
\end{align*}

\textbf{Latent Density:} The log-density of $\mathbf{z} \sim \mathcal{N}(\mathbf{0}, \mathbf{I})$ is:
\[
\log p(\mathbf{z}) = -\frac{1}{2} \left( \|\mathbf{z}\|^2 + D \log(2\pi) \right).
\]
Combining the two components, the final expression for the conditional log-likelihood is:
\[
\log p(\mathbf{x} \mid \mathbf{y}) = \log p(\mathbf{z}) + \sum_{j=1}^{D} \log s_j(\mathbf{y}),
\]
where $\mathbf{z} = f_\psi^{-1}(\mathbf{x}, \mathbf{y}) = \left( \mathbf{x} - t(\mathbf{y}) \right) \oslash \exp(\log s(\mathbf{y}))$.

\begin{acknowledgement}
The AI-TranspWood project, HORIZON-CL4–2023-RESILIENCE- 01–23 (Grant Agreement 101138191), is gratefully acknowledged. This project is co-funded by the European Union. Views and opinions expressed are however those of the author(s) only and do not necessarily reflect those of the European Union or HaDEA. Neither the European Union nor the granting authority can be held responsible for them.
\end{acknowledgement}

\bibliography{sample}

\begin{thebibliography}{10}
\providecommand{\url}[1]{#1}
\csname url@samestyle\endcsname
\providecommand{\newblock}{\relax}
\providecommand{\bibinfo}[2]{#2}
\providecommand{\BIBentrySTDinterwordspacing}{\spaceskip=0pt\relax}
\providecommand{\BIBentryALTinterwordstretchfactor}{4}
\providecommand{\BIBentryALTinterwordspacing}{\spaceskip=\fontdimen2\font plus
\BIBentryALTinterwordstretchfactor\fontdimen3\font minus \fontdimen4\font\relax}
\providecommand{\BIBforeignlanguage}[2]{{%
\expandafter\ifx\csname l@#1\endcsname\relax
\typeout{** WARNING: IEEEtran.bst: No hyphenation pattern has been}%
\typeout{** loaded for the language `#1'. Using the pattern for}%
\typeout{** the default language instead.}%
\else
\language=\csname l@#1\endcsname
\fi
#2}}
\providecommand{\BIBdecl}{\relax}
\BIBdecl

\bibitem{azarkish2019reliability}
H.~Azarkish and M.~Rashki, ``Reliability and reliability-based sensitivity analysis of shell and tube heat exchangers using monte carlo simulation,'' \emph{Appl. Therm. Eng.}, vol. 159, p. 113842, 2019.

\bibitem{doronin2011online}
A.~Doronin and I.~Meglinski, ``Online object oriented monte carlo computational tool for the needs of biomedical optics,'' \emph{Biomed. Opt. Express}, vol.~2, no.~9, pp. 2461--2469, 2011.

\bibitem{seyedheydari2022electromagnetic}
F.~Seyedheydari, K.~Conley, P.~Yl{\"a}-Oijala, A.~Sihvola, and T.~Ala-Nissila, ``Electromagnetic response and optical properties of anisotropic cusbs2 nanoparticles,'' \emph{JOSA B}, vol.~39, no.~7, pp. 1743--1751, 2022.

\bibitem{wang1995mcml}
L.~Wang, S.~L. Jacques, and L.~Zheng, ``M{CML}-{M}onte {C}arlo modeling of light transport in multi-layered tissues,'' \emph{Comput. Methods Programs Biomed.}, vol.~47, no.~2, pp. 131--146, 1995.

\bibitem{conley2021silica}
K.~Conley, S.~Moosakhani, V.~Thakore, Y.~Ge, J.~Lehtonen, M.~Karttunen, S.-P. Hannula, and T.~Ala-Nissila, ``Silica-silicon composites for near-infrared reflection: A comprehensive computational and experimental study,'' \emph{Ceram. Int.}, vol.~47, no.~12, pp. 16\,833--16\,840, 2021.

\bibitem{lu2019efficient}
D.~Lu and D.~Ricciuto, ``Efficient surrogate modeling methods for large-scale earth system models based on machine-learning techniques,'' \emph{Geosci. Model Dev.}, vol.~12, no.~5, pp. 1791--1807, 2019.

\bibitem{kaya2018surrogate}
M.~Kaya and S.~Hajimirza, ``Surrogate based modeling and optimization of plasmonic thin film organic solar cells,'' \emph{Int. J. Heat Mass Transf.}, vol. 118, pp. 1128--1142, 2018.

\bibitem{goodfellow2016deep}
I.~Goodfellow, \emph{Deep learning}.\hskip 1em plus 0.5em minus 0.4em\relax MIT press, 2016, vol. 196.

\bibitem{gal2016dropout}
Y.~Gal and Z.~Ghahramani, ``Dropout as a bayesian approximation: Representing model uncertainty in deep learning,'' in \emph{international conference on machine learning}.\hskip 1em plus 0.5em minus 0.4em\relax PMLR, 2016, pp. 1050--1059.

\bibitem{rezende2015variational}
D.~Rezende and S.~Mohamed, ``Variational inference with normalizing flows,'' in \emph{International conference on machine learning}.\hskip 1em plus 0.5em minus 0.4em\relax PMLR, 2015, pp. 1530--1538.

\bibitem{kobyzev2020normalizing}
I.~Kobyzev, S.~J. Prince, and M.~A. Brubaker, ``Normalizing flows: An introduction and review of current methods,'' \emph{IEEE Trans. Pattern Anal. Mach. Intell.}, vol.~43, no.~11, pp. 3964--3979, 2020.

\bibitem{park2024solving}
J.~H. Park, J.~Lee, and J.~Hwang, ``Solving inverse problems using normalizing flow prior: Application to optical spectra,'' \emph{Phys. Rev. B}, vol. 109, no.~16, p. 165130, 2024.

\bibitem{durkan2019neural}
C.~Durkan, A.~Bekasov, I.~Murray, and G.~Papamakarios, ``Neural spline flows,'' \emph{NeurIPS}, vol.~32, 2019.

\end{thebibliography}

\bibliographystyle{IEEEtran}

\end{document}